\documentclass[10pt,twocolumn,letterpaper]{article}

\usepackage{cvpr}
\usepackage{times}
\usepackage{epsfig}
\usepackage{graphicx}
\usepackage{amsmath}
\usepackage{amssymb}
\graphicspath{{./images/}}
\usepackage{multirow}
\usepackage{algorithm}
\usepackage{algpseudocode}
\DeclareMathOperator*{\argmin } {min}

\usepackage[pagebackref=true,breaklinks=true,letterpaper=true,colorlinks,bookmarks=false]{hyperref}

\cvprfinalcopy 

\def\cvprPaperID{4149} 

\ifcvprfinal\fi
\begin{document}

\title{Learning Structure and Strength of CNN Filters for Small Sample Size Training}

\author{Rohit Keshari, Mayank Vatsa, Richa Singh \\
 IIIT-Delhi, India\\
{\tt\small \{rohitk, mayank, rsingh\}@iiitd.ac.in}
\and
Afzel Noore \\
Texas A\&M University-Kingsville, USA\\
{\tt\small Afzel.Noore@tamuk.edu}
}

\maketitle

\begin{abstract}

Convolutional Neural Networks have provided state-of-the-art results in several computer vision problems. However, due to a large number of parameters in CNNs, they require a large number of training samples which is a limiting factor for small sample size problems. To address this limitation, we propose SSF-CNN which focuses on learning the ``structure" and ``strength" of filters. The structure of the filter is initialized using a dictionary based filter learning algorithm and the strength of the filter is learned using the small sample training data. The architecture provides the flexibility of training with both small and large training databases, and yields good accuracies even with small size training data. The effectiveness of the algorithm is first demonstrated on MNIST, CIFAR10, and NORB databases, with varying number of training samples. The results show that SSF-CNN significantly reduces the number of parameters required for training while providing high accuracies on the test databases. On small sample size problems such as newborn face recognition and Omniglot, it yields state-of-the-art results. Specifically, on the IIITD Newborn Face Database, the results demonstrate improvement in rank-1 identification accuracy by at least 10\%.


\end{abstract}

\section{Introduction}

\begin{figure}[!t]
	\centering
	\includegraphics[width=.5\textwidth]{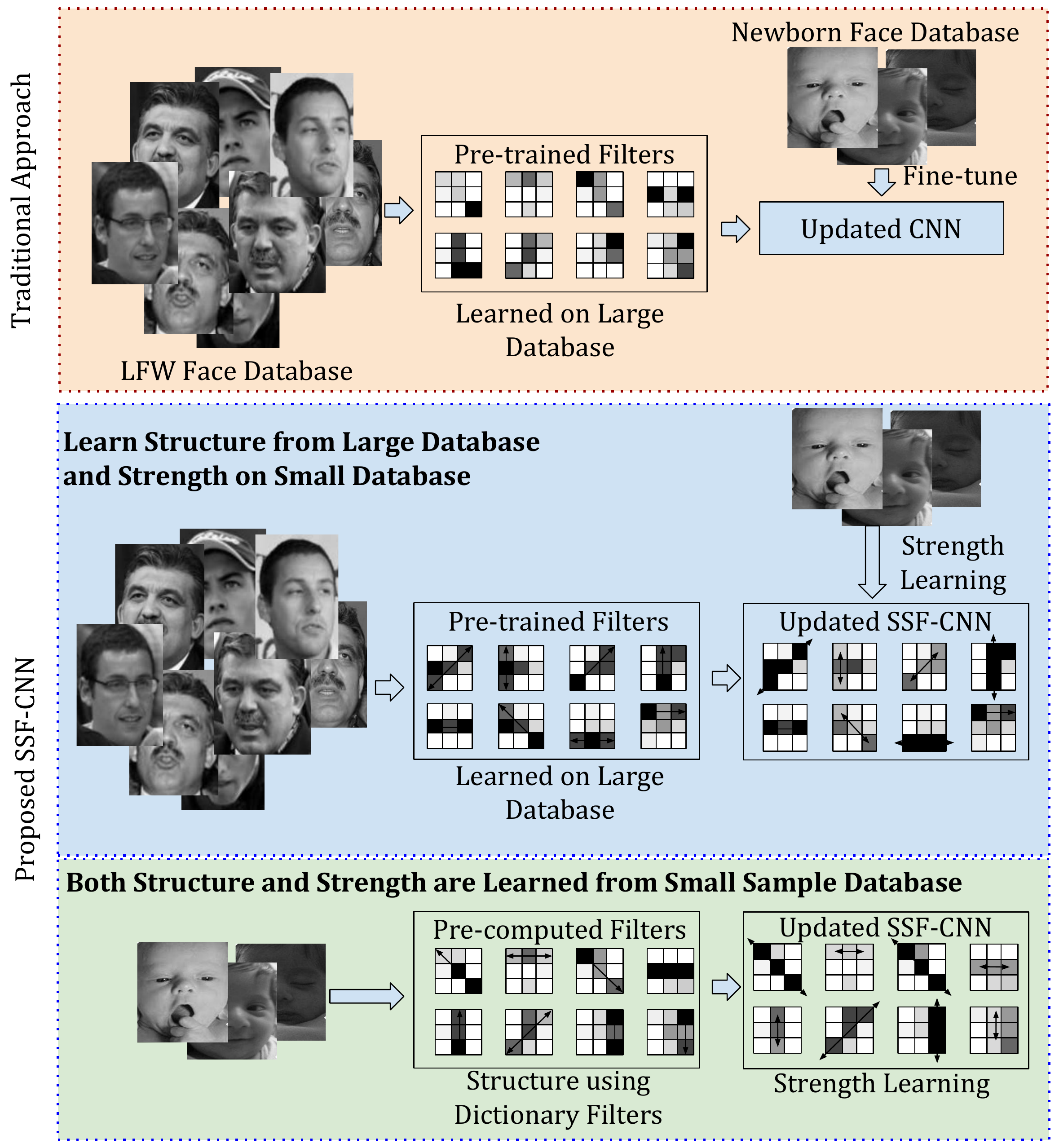}
	\caption{Face recognition models trained on adult face images may not provide good performance for newborn face recognition. SSF-CNN proposes to learn structure and strength of the filters for improving the classification performance for small sample databases.}	
	\label{fig:intro} 	
\end{figure}

Convolutional Neural Network (CNN) is a multilayer representation learning architecture which has received immense success in multiple applications such as object classification, image segmentation, and natural language processing. From LeNet~\cite{lecun1989backpropagation} to AlexNet~\cite{krizhevsky2012imagenet}, GoogleNet~\cite{szegedy2015going}, VGG-Net~\cite{simonyan2014very}, ResNet~\cite{he2016deep}, and now DenseNet~\cite{huang2017densely}, given large training data, CNNs have shown state-of-the-art performance for several applications. However, large training data is also a limiting requirement for applications with small sample size and many of these architectures easily overfit on small training samples. For example, as shown in Figure \ref{fig:intro}, a face recognition model trained on large training data of adult faces (e.g. CelebA or LFW databases) may not provide good performance when tested for newborn face recognition \cite{newbornbtas, bharadwaj2016domain}. In newborn face recognition, the available training data may be small and therefore, even after fine-tuning, standard deep learning based face recognition models may not yield high performance.

To address the challenge of small sample size, researchers have proposed algorithms focusing on CNN initialization tricks and modifications to CNN architecture. Erhan~\textit{et al.} \cite{erhan2010does} have investigated the importance of unsupervised pre-training of deep architecture and empirically shown that pre-trained weights of the network generalize better than randomly initialized weights. Similarly, Mishkin and Matas \cite{mishkin2015all} have proposed Layer-Sequential Unit-Variance (LSUV) initialization that utilizes the orthonormal matrices to initialize the weights of each convolutional layer and normalize the weight to the unit variance. Along the same lines, pre-defined handcrafted filters are also proposed to handle the small sample size problem. For example, And{\'e}n and Mallat \cite{anden2011multiscale} propose Scattering network (ScatNet) which is a CNN like architecture where pre-defined Morlet filter bank is utilized to extract features. However, these handcrafted filters may not represent the true distribution of the data and hence extract not-so-meaningful features. To overcome this limitation, Oyallon \textit{et. al.} \cite{oyallon2017scaling} have proposed hybrid network, where they have utilized ScatNet feature followed by CNN architecture. Similarly, Chan~\textit{et. al.}~\cite{chan2015pcanet} propose PCANet architecture that utilizes Principal Component Analysis (PCA) to learn the filter banks. They also present an extension, termed as LDANet, in which the selection of the cascade filters are trained from Linear Discriminant Analysis (LDA). Gan~\textit{et al.}~\cite{gan2015pca} propose a PCA-based Convolutional Network (PCN) which has the influence of both CNN~\cite{jarrett2009best} and PCANet~\cite{chan2015pcanet}. Dan~\textit{et al.}~\cite{wu2015kernel} utilize the concept of kernel PCA to further improve the PCANet architecture. Zeng~\textit{et al.}~\cite{zeng2015tensor} propose a multilinear discriminant analysis network (MLDANet) which is a variant of PCANet and LDANet. Feng~\textit{et al.}~\cite{feng2015dlanet} propose Discriminative Locality Alignment Network (DLANet) which is based on manifold learning. These architectures learn filters in stack-wise manner, and once the network (filters) is trained, generally, it is not allowed to fine-tune the filters on other databases.

In other research directions for small sample size training, Mao~\textit{et al.}~\cite{mao2006new} propose a neural network learning method based on posterior probability (PPNN) to improve the accuracy. Ngiam~\textit{et al.}~\cite{ngiam2010tiled} propose tied weights in a filter using tiling parameter which handles the total number of learning parameters. In another work, Indian Buffet Process (IBP) priors are utilized to propose semi-supervised ibpCNN which shows better generalizability \cite{feng2015learning}. Xiong~\textit{et al.}~\cite{xiong2016regularizing} propose Structured Decorrelation Constrained (SDC) for hidden layers. The authors have also proposed a novel approach termed as Regularized Convolutional Layers (Reg-Conv) that can help SDC to regularize the complex convolutional layers. Similarly, Cogswell~\textit{et al.}~\cite{cogswell2015reducing} propose DeConv loss for CNN architecture that helps in training small databases.

One of the major problems with adapting pre-trained CNN models for small sample size problems, as mentioned previously, is large amount of parameters; therefore, insufficient training samples may cause overfitting. If we reduce these parameters to a significantly small number, then the problem can be addressed in a better way. This paper focuses on two novel ways to develop CNN based feature representation algorithm for small sample size problems: (i) associating ``strength'' parameter to control the effect of each pre-trained filter, and (ii) utilizing a generalizable approach that pre-learns the ``structure" of the filters using small training samples. The proposed architecture is motivated from ScatNet but in place of pre-defined filters, we utilize dictionary learning model to \textit{pre-learn} the filters. Further, unlike CNN approaches where we update the weights in every iteration, we introduce strength of the filter and update only the strength parameter not the filters. The introduction of ``\textit{strength}'' of filters significantly reduces the number of parameters to learn (detailed calculations shown later) and therefore avoids overfitting with limited training. Experiments are performed on object classification databases, MNIST \cite{lecun1998gradient}, CIFAR-10 \cite{krizhevsky2009learning}, NORB \cite{lecun2004learning}, Omniglot \cite{lake2011one}, and a challenging small sample size database of newborn faces \cite{bharadwaj2016domain}. Comparison with existing algorithms show that the proposed approach achieves state-of-the-art performance for small sample size problems and significantly reduces the number of parameters to learn/fine-tune.   

\section{Proposed SSF-CNN}

\begin{figure}
	\centering
	\includegraphics[width=.5\textwidth]{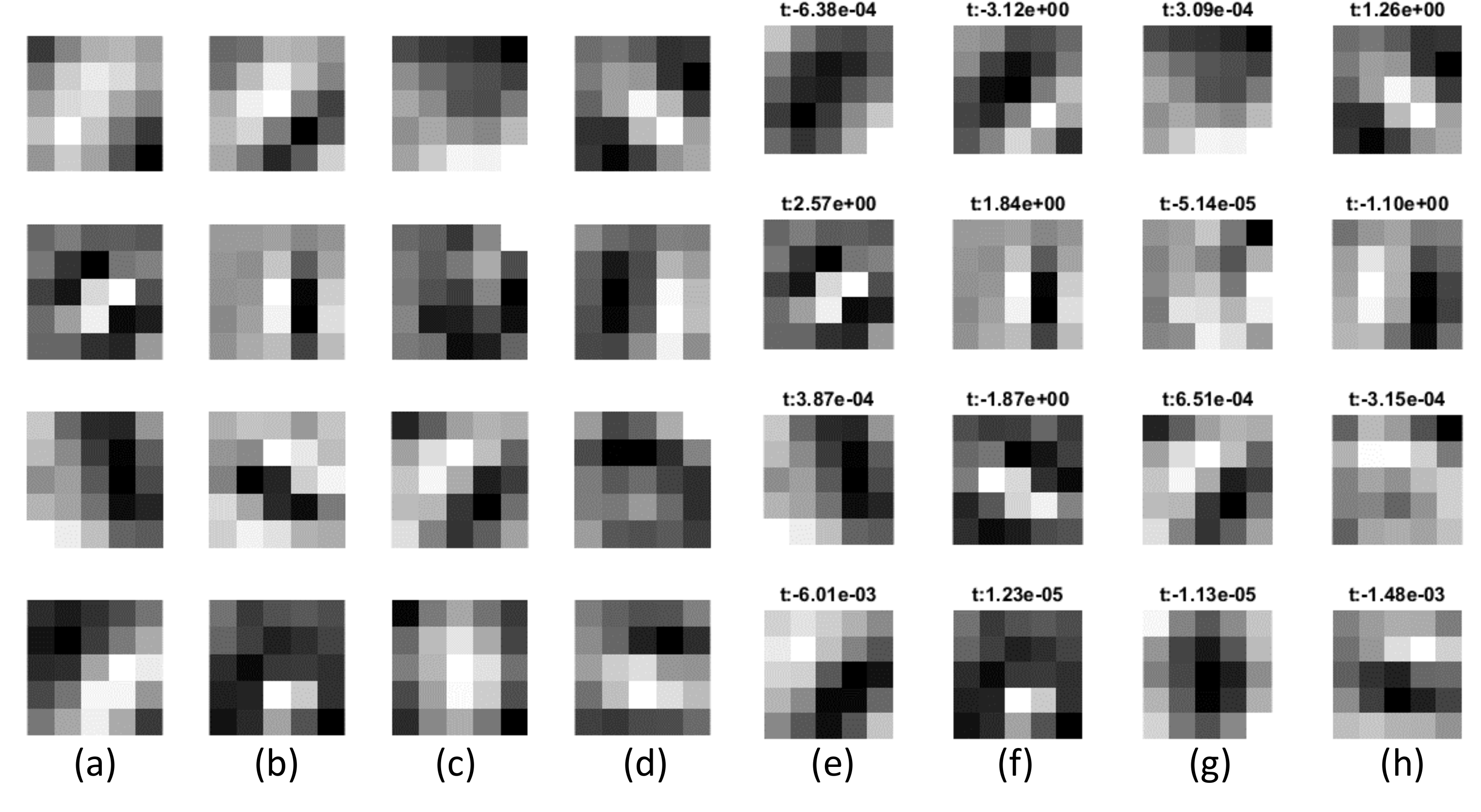}
	\caption{Filters (a) to (d) are dictionary trained filters. Filters (e) to (h) illustrate the change due to the proposed strength parameter in CNN architecture. These filters are trained on MNIST database.}	
	\label{fig:t_learning} 	
\end{figure}

It is difficult to learn the entire network from scratch while training with small size databases. Existing approaches with pre-defined or handcrafted filters \cite{anden2011multiscale}, and pre-trained filters \cite{chan2015pcanet, gan2015pca, zeng2015tensor}, may not allow fine-tuning the filters and therefore, the learned model may not represent the true data distribution for small sample size problems. To mitigate these challenges, we propose a novel approach, termed as Structure and Strength Filtered CNN (SSF-CNN), which has two components: (i) structure of the filter and (ii) strength of the filter. It is our hypothesis that structure of the CNN filters can be learned from either domain specific larger databases or from other representation learning paradigms that require less training data for instance, dictionary learning \cite{dictionary3, dictionary1}. It is well known that matrix factorization or dictionary learning allows us to learn the \textit{dictionary} that helps encoding the representative features. If we represent CNN filters using dictionary, it can provide the ``structure''; however, it may not be well optimized for the classification task. Therefore, the next part of the framework is computing ``strength" of every filter to adapt the weights of these filters according to the data characteristics. Strength can be interpreted as the attuning parameter to update or adapt the filters based on the small size training data. For illustration, columns (a) to (d) in Figure~\ref{fig:t_learning} represent the samples from trained dictionary filters for the MNIST database and columns (e) to (h) represent the updated filters where changes are due to the strength parameter. 



Formally, in the proposed approach, first the hierarchical dictionary filters are learned to initialize the CNN, followed by learning the strength parameter to train the CNN model. We introduce strength parameter `$\mathbf{t}$' for the CNN filters `$\mathbf{W}$' which allows the network to assign \textit{weight} for each filter based on its structural importance. In CNN model, strength and structural parameters $\mathbf{t}$ and $\mathbf{W}$ can be learned in two ways: 1) pre-train $\mathbf{W}$, use it in CNN by freezing the values of $\mathbf{W}$ followed by learning the strength $\mathbf{t}$, and 2) pre-train $\mathbf{W}$ which is used to initialize the CNN model followed by learning $\mathbf{t}$ and $\mathbf{W}$ iteratively. While the second approach which simultaneously learns both structure and strength may be desirable, the first approach requires very few parameters to be trained in CNN model. We next describe the approach to hierarchically learn $\mathbf{W}$, filters of CNN model, using dictionary learning followed by learning the strength parameter $\mathbf{t}$.

\begin{figure}
	\centering
	\includegraphics[width=0.5\textwidth]{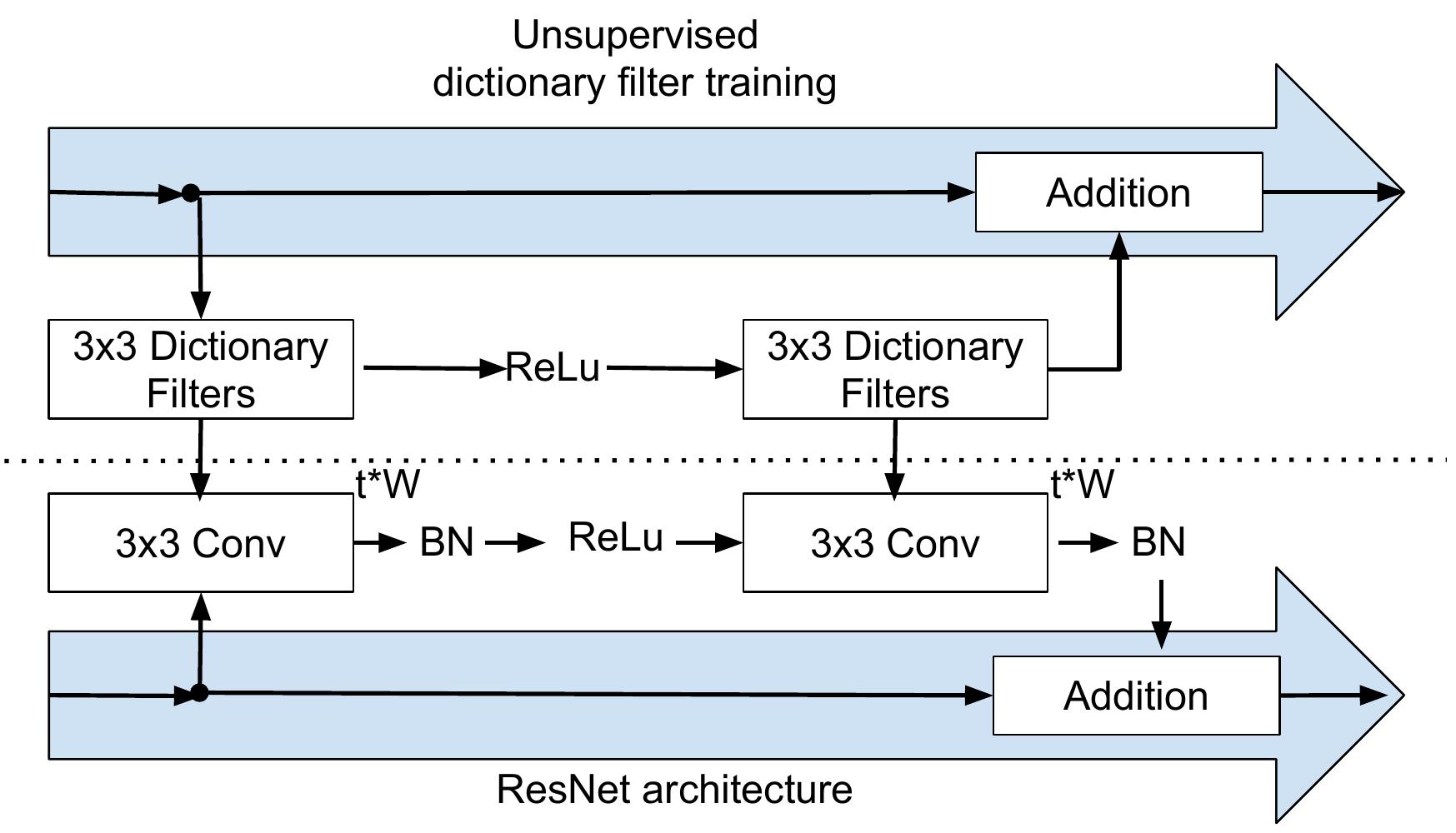}
	\caption{The proposed SSF-CNN architecture for initializing the ResNet architecture with the filters learned from dictionary.}	
	\label{fig:dictionary_RNet} 	
\end{figure}

\begin{algorithm}[t]

	\caption{Hierarchical Dictionary Filter Learning}

	\label{DictAlgo}

	\begin{algorithmic}[1]

		\State \textbf{Notation:} $N$ is a number of training samples, $n$ number of extracted patches, $y$ is a patch from $Y$

		\State \textbf{Input:} $X_N$

		\State \textbf{Output:}	$D$		

		\For{each layer $l:=1$ to $numLayer$ }

		\State $[x^n]^N \gets extractPatch(X_N)$

		\State Y $\gets$ reshape($[x^n]^N$)

		\State $\min_{D \in \mathbb{R}^{m\times k}} \frac{1}{n}\sum^n_{i=1} \min_{\alpha^i } (\frac{1}{2}||y^i-D^l\alpha^i||^2_2+\lambda||\alpha^i||_{1})$

		\State W $\gets$ reshape($D^l$)

		\For{$j:=1$ to $N$}

		\State $fmap_j=X_j*W$

		\EndFor

		\State $X_N\gets ReLu(fmap)$

		\EndFor			

	\end{algorithmic}

\end{algorithm}

%
%

\subsection{Learning Structure of Filters}


In this research, we propose to use dictionary learning algorithm for learning the structure of the filters. The algorithm can be divided into two steps: 1) learn hierarchical dictionary filters and utilize trained dictionary filters to initialize the CNN, and (2) train CNN with dictionary initialized filters.

\vspace{6pt}
\noindent \textbf{Hierarchical Dictionary Filter Learning}: Dictionary learning focuses on learning a sparse representation of the input data in the form of a linear combination of basic elements or atoms \cite{dictionary4, lee99, mairal2010online, dictionary3, dictionary1}. For a given input $\mathbf{Y}$, a dictionary $\mathbf{D}$ is learned along with the coefficients $\mathbf{\alpha}$:
\begin{equation} \label{KSVD}
{
\argmin_{\textit{$\mathbf{D, \alpha}$}} \left \|\mathbf{Y} - \mathbf{D\alpha} \right \|_{F}^{2},\ such\ that \ \left \|\mathbf{\alpha}\right \|_{0} \leq \tau
}
\end{equation}

\noindent where, the $\ell_0$-norm imposes a constraint of sparsity on the learned coefficients and $\tau$ corresponds to the maximum number of non-zero elements. Often, the $\ell_0$-norm is relaxed and the updated dictionary learning formulation can be written as:
\begin{equation}
\label{eq:dict}
\argmin_{\textit{$\mathbf{D, \alpha}$}} \left \|\mathbf{Y} - \mathbf{D\alpha} \right \|_{F}^{2} + \lambda||\mathbf{\alpha}||_{1}
\end{equation}

\noindent where, $\lambda$ is a regularization parameter which controls the sparsity promoting $\ell_1$-norm. In this research, we utilize dictionary learning to pre-train the filters of CNN in a hierarchical manner. As shown in Algorithm~\ref{DictAlgo}, a hierarchical dictionary learning technique is utilized to initialize the CNN model (ResNet~\cite{he2016deep}). The trained dictionary atoms are used to convolve over the input image. After convolution, feature maps are normalized according to the activation function (e.g. ReLu) used in CNN models. Figure \ref{fig:dictionary_RNet} presents the structure of a block of the SSF-ResNet architecture. The extracted feature map is an input for the next level of the hierarchical dictionary. In this manner, the number of dictionary layers is same as the number of convolutional layers in CNN models. In Algorithm~\ref{DictAlgo} $extractPatch$ function is used to tessellate the input image into small patches. The trained dictionary is organized in the two-dimensional array where each filter is arranged in one column. These learned filters are reshaped and convolved over the input image to produce the feature maps for the next level of the dictionary.

\begin{figure}[t]
	\centering
	\includegraphics[width=0.485\textwidth]{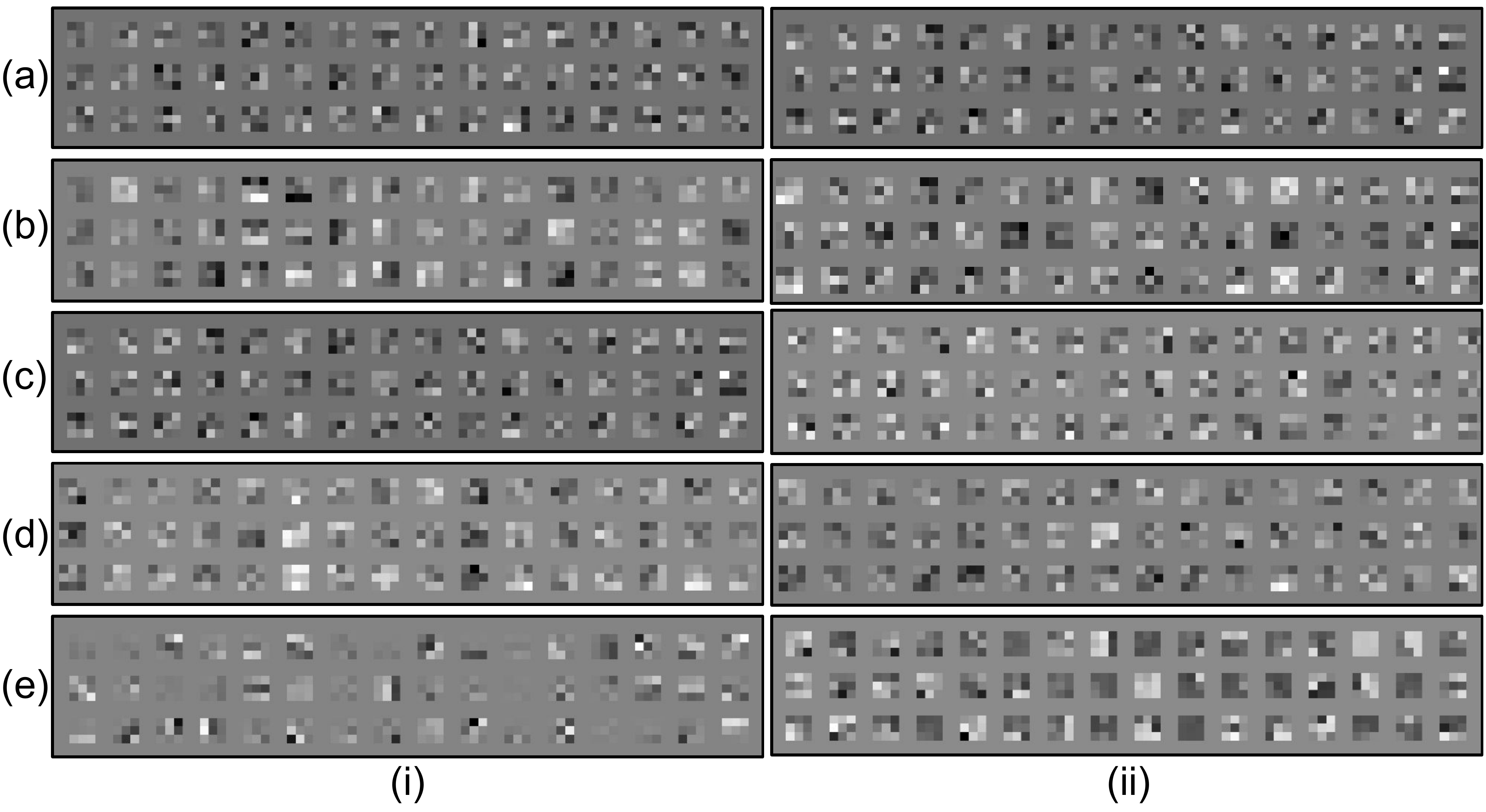}
	\caption{Filter visualization of the (i) $1^{st}$ layer and (ii) $2^{nd}$ layer of the ResNet architecture on CIFAR10 dataset. (a) Xavier~\cite{glorot2010understanding} initialized filters at zero epoch, (b) Xavier~\cite{glorot2010understanding} initialized filters are trained on 1000 training samples, (c) MSRA~\cite{he2015delving} initialized filters at zero epoch, (d) MSRA~\cite{he2015delving} initialized filters are trained on 1000 training samples, and (e) Dictionary initialized filters at zero epoch. For better visualization, only 16 filters are used from the $2^{nd}$ layer.}	
	\label{fig:cifar10_dict} 	
\end{figure}

\vspace{6pt}

\noindent \textbf{Training CNN with Dictionary Initialized Filters}: Typically, CNN has multiple convolutional layers, each layer has multiple filters, and these filters are trained using stochastic gradient descent (SGD)~\cite{lecun1988theoretical}.  
For input $\mathbf{X}$ and convolutional filter $\mathbf{W}$, the convolutional function of the CNN can be defined as $f(\mathbf{X},\mathbf{W},b)=\mathbf{X}*\mathbf{W}+b$, where $*$ is the convolutional operation and $b$ is the bias. A CNN architecture is designed by stacking multiple convolutional and pooling layers. These deep CNN architectures are trained in two passes: 1) forward pass and 2) backward pass. In the forward pass, network propagates the input signal to the last classification layer. In backward pass, the error $\delta_j^l$ for each layer $l$ on node $j$ is computed with respect to the cost and the weights of the CNN filters are updated accordingly.

Let $\mathbf{a}^l$ be the output feature map at $l^{th}$ layer of the CNN with a cost function $C$. The weights are updated as per the gradient direction, i.e. $\Delta \mathbf{W}^l=\frac{\partial C}{\partial \mathbf{W}^l}$. Using chain rule, $\Delta \mathbf{W}^l=\mathbf{a}^{l-1}\mathbf{\delta}^l$. In traditional CNN learning, the weights are initialized in different ways such as Xavier~\cite{glorot2010understanding}, or MSRA~\cite{he2015delving} approach and even randomly. In this research, we propose initialization of the CNN filters using dictionary learned filters as discussed above. As shown in Figure \ref{fig:cifar10_dict}, filters learned from the dictionary learning technique show more ``structure'' than traditional approaches, particularly with small training data. While dictionary initialization helps in finding improved features, updating the filters in a traditional manner still requires large parameter space, which is not conducive for small training data. In the next subsection, we present the proposed approach of incorporating strength of the filters and not update the filters using SGD which reduces the number of learning parameters significantly.  

\begin{figure}[!t]
	\centering
	\includegraphics[width=.5\textwidth]{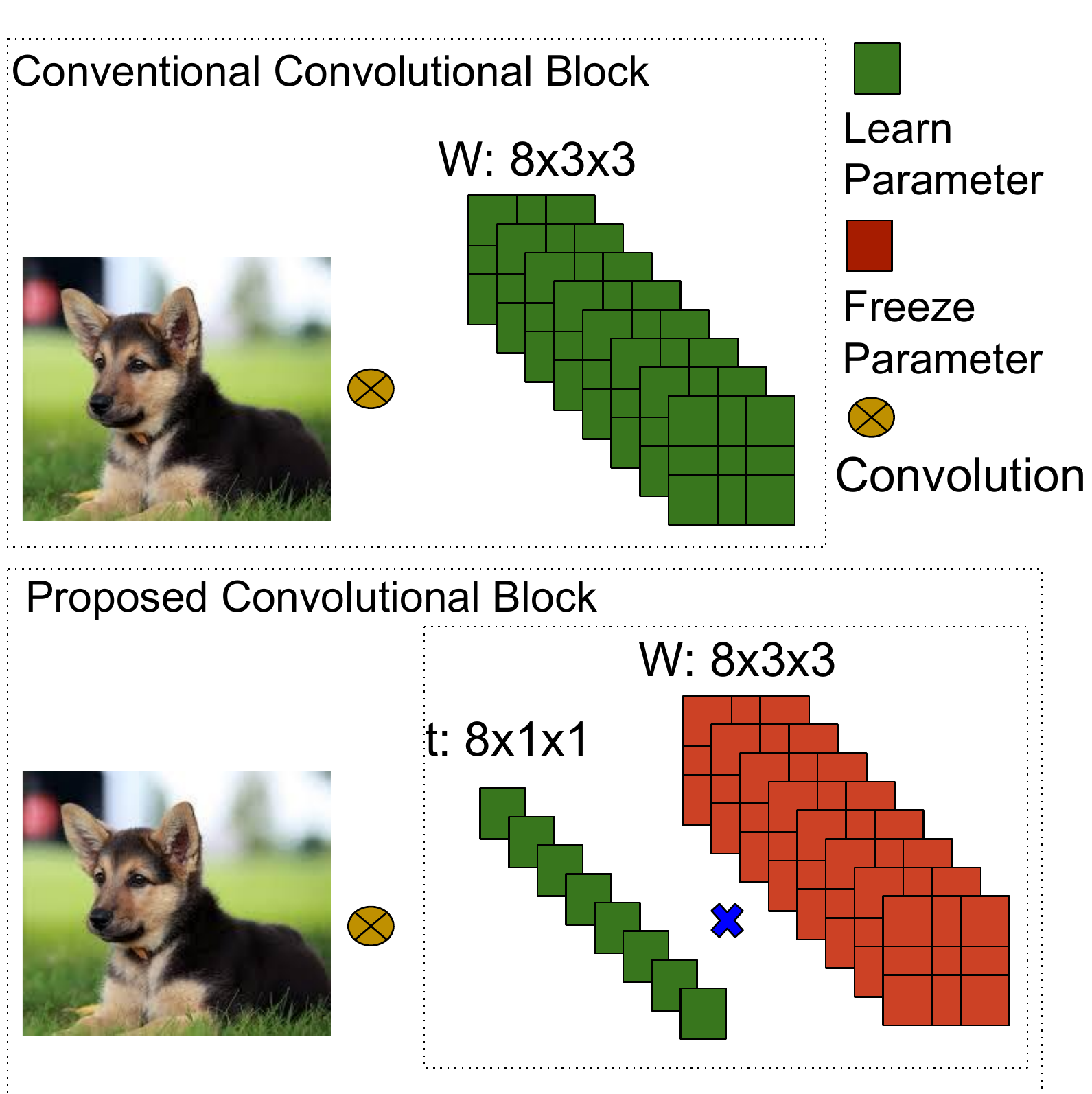}
	\caption{Illustrating the concept of learning the strength of a filter which significantly reduces the number of training parameters.}	
	\label{fig:tCNN} 	
\end{figure}

\subsection{Learning Filter Strength}
The proposed concept of learning strength of the filter is illustrated in Figure~\ref{fig:tCNN}. Here, we freeze the values of filters obtained from dictionary learning technique and update only the strength of the filter. As shown in Figure~\ref{fig:tCNN}, this significantly reduces the number of learning parameters. For $l^{th}$ layer, the strength parameter  `$\mathbf{t}^{l}$' is learned using stochastic gradient descent method; i.e. a scalar value is learned rather than learning the complete filter. The proposed process can be written as, 

\begin{equation}
\label{eq:propoedF}
f(\mathbf{X},\mathbf{W},b,\mathbf{t})=\mathbf{X}*(\mathbf{t \odot W})+b
\end{equation}

\noindent where, $(\mathbf{t \odot W})$ represents element-wise multiplication. The pre-trained filters learned from dictionary learning or pre-trained model are selected and the only variable to be learned is $\mathbf{t}$ which can be learned using SGD. Since $|\mathbf{W}|>>|\mathbf{t}|$, even small training data can be used to train the network. In literature, various regularization techniques have been utilized for better convergence. Existing regularization techniques such as dropconnect and $\ell_1$ regularization can also be used while learning $\mathbf{t}$. 

\section{Experimental Results}

The effectiveness of the proposed algorithm is evaluated on multiple databases with state-of-the-art CNN architectures including ResNet \cite{he2016deep} and DenseNet \cite{huang2017densely}. The details of experiments and results are described below.

\subsection{Database and Experimental Protocol}

Since the proposed architecture is for small size training data, the experiments are performed with varying training sizes on three databases: MNIST~\cite{lecun1998gradient}, CIFAR10~\cite{krizhevsky2009learning}, and NORB~\cite{lecun2004learning}. More specifically, as shown in Table \ref{tb:DB}, the experiments are performed with 14 data sizes, $100, 200,\cdots, 1000, 2000,\cdots, 5000$. The proposed algorithm is also tested with the complete/standard training set. Further, experiments are performed on an interesting and small sample size problem of newborn face recognition~\cite{bharadwaj2016domain}. The newborn database has images from 96 babies and as per the predefined protocol \cite{bharadwaj2016domain}, training data consists of images from 10 newborns and the remaining images, corresponding to 86 newborns, are used for testing (with 1, 2, 3, and 4 images per subject in the gallery). Finally, experiments are also performed on the Omniglot database \cite{lake2011one} which comprises $1623$ handwritten characters pertaining to $50$ different alphabets. The background database has 30 alphabets and evaluation set has 20 alphabets. All the experiments are performed with five fold cross validation and average accuracies are reported in next subsections. 


\begin{table}[!]
\centering
\footnotesize
\caption{Experimental protocols for MNIST, CIFAR-10 and NORB databases.}
\label{tb:DB}
\begin{tabular}{|l|c|c|c|c|c|l|c|}
\hline
\multirow{2}{*}{\textbf{Databases}} & \multirow{2}{*}{\textbf{Small Training Data}} & \multirow{2}{*}{\textbf{\begin{tabular}[c]{@{}l@{}}Standard \\ Training\end{tabular}}} & \multirow{2}{*}{\textbf{\begin{tabular}[c]{@{}c@{}}Standard \\ Testing\end{tabular}}} \\
                                    &      &           &                                    \\ \hline
\textbf{MNIST}   & $100:100:1k; 1k:1k:5k$         & \multicolumn{1}{c|}{50k}         & 10k    \\ \hline
\textbf{CIFAR-10}   & $100:100:1k; 1k:1k:5k$         & \multicolumn{1}{c|}{40k}      & 10k     \\ \hline
\textbf{NORB}       & $100:100:1k;1k:1k:5k$        & \multicolumn{1}{c|}{20k}       & 24.3k \\ \hline

\end{tabular}
\end{table}

\subsection{Implementation Details}

\begin{figure}[!t]
	\centering
	\includegraphics[width=0.5\textwidth]{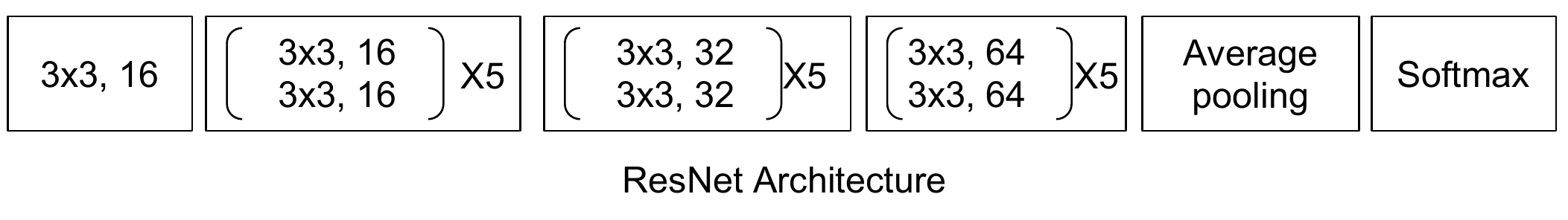}
	\caption{Illustrating the ResNet architecture used in the experiments.}
	\label{fig:arch}
\end{figure}

To demonstrate the results of the proposed SSF-CNN, a popular ResNet~\cite{he2016deep} architecture is used. Figure~\ref{fig:arch} illustrates the ResNet architecture which has 1 input layer, 31 convolutional layers, 1 global pooling layer, and 1 softmax layer. The strength parameter is regularized with both ElasticNet \cite{zou2005regularization} ($\lambda_1 |\textbf{t}|_1+\lambda_2 |\textbf{t}|_2$) and DropConnect ~\cite{wan2013reg}. It is experimentally observed that in the first $20$ epochs, $\lambda_2$ is $0.0001$ and $\lambda_1$ is $0$. After $20$ epochs both the regularization constants are set to $0.0001$. $\ell_1$ regularization introduces sparsity in $\textbf{t}$ parameters and helps to fadeout the less contributing filters thus improving the strength of filters with large contribution. Further, at every epoch, dropconnect parameter is randomly initialized by $Bernoulli(pr)$ where $pr$ has $0.8$ and $0.2$ probability for generating $1s$ and $0s$ respectively.

The proposed model utilizes a dictionary and pre-trained model to initialize and train the CNN filters.  Specifically, dictionary filters are learned using K-SVD algorithm~\ref{DictAlgo}. These dictionaries are layered in a similar manner as CNN layers and are referred to as hierarchical dictionary. The parameter values for K-SVD such as sparsity parameter, the total number of iteration, and batch size for dictionary have been initialized with $0.1$, $1000$, and $100$ respectively. The input signal for dictionary are patches extracted from randomly selected $N$ number of balanced training samples. The value of $N$ varies from $100, 200,\cdots,1000, 2000,\cdots, 5000$, as shown in Table \ref{tb:DB}.

\subsection{Parameter Learning}

In traditional ResNet architecture, total number of parameters to be learned in convolutional layers for the CIFAR-10 dataset is $242,352$. 
 On the other hand, in the proposed SSF-CNN, total number of strength parameters to be learned for the same database is $26,928$. This shows that the proposed architecture reduces the total number of parameters to be learned by ${1/9}^{th}$ factor in each convolutional layer. Similarly, for other databases and architectures, we observe reduced number of parameters to train.

\subsection{Results on Limited Training Data - MNIST, CIFAR-10, and NORB}

\begin{figure*}[!t]
	\centering
	\includegraphics[width=\textwidth]{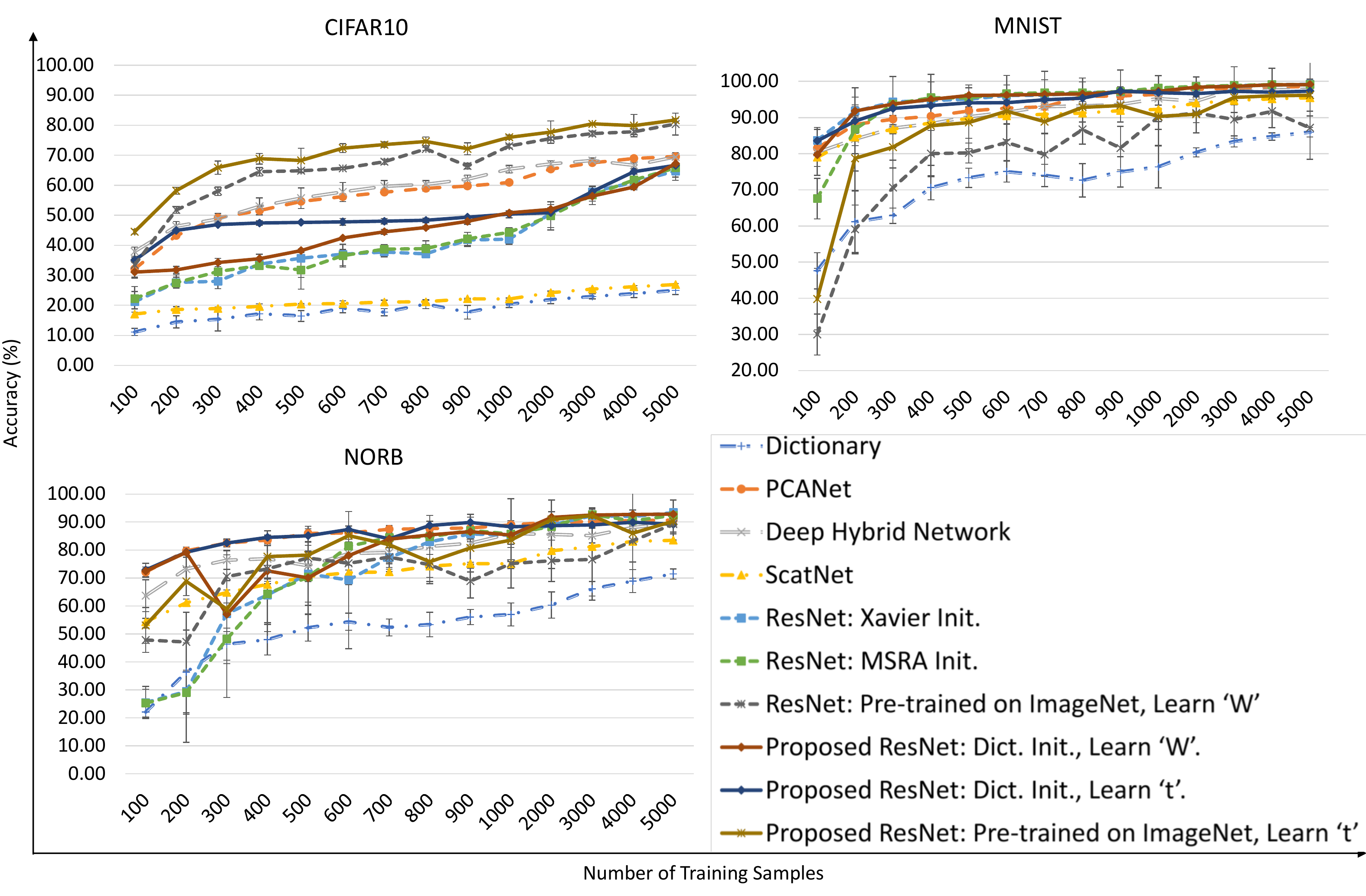}
	\caption{Classification accuracies (\%) for CIFAR-10, MNIST, and NORB databases with varying the number of training samples.}
	\label{fg:db_graphs} 	
\end{figure*}

The main focus of the proposed SSF-CNN is to learn the deep neural network models with a small number of training samples. 
Since the proposed initialization is performed using dictionary learning, we also compute the results of shallow dictionary which serves as the baseline for all the experiments. We have also compared the proposed algorithm with PCANet~\cite{chan2015pcanet}, Deep Hybrid Network~\cite{oyallon2017scaling}, ScatNet~\cite{anden2011multiscale}, ResNet initialized with Xavier~\cite{glorot2010understanding}, and ResNet initialized with MSRA~\cite{he2015delving}. For the proposed SSF-CNN, two sets of results are computed based on the manner in which the parameters $\mathbf{W}$ and $\mathbf{t}$ are learned.

\begin{itemize}
	\item \textbf{Experiment 1} - \textit{Learn $\mathbf{W}$}: Initialized filters are fine-tuned while doing backpropagation.

\item \textbf{Experiment 2} - \textit{Learn $\mathbf{t}$, Freeze $\mathbf{W}$}: Only the strength parameter $\mathbf{t}$ is learned while the initialized filters are not updated.	 
\end{itemize}

\noindent \textbf{Filter Visualization:} We first analyze the filters learned from the proposed method and CNN. Figure \ref{fig:cifar10_dict} shows the first and second layer filters trained on CIFAR-10 database: (a) \& (c) showcase filters with two existing initialization techniques in CNN architecture, (b) \& (d) trained CNN filters on $1000$ training samples, and (e) trained dictionary filters on $1000$ training samples. In Figure \ref{fig:cifar10_dict}, it can be observed that dictionary trained filters have less noisy patterns compare to CNN trained filters on small data. In literature, Zeiler and Fergus \cite{zeiler2014visualizing} have also suggested that the filters that have structural properties are good while the ones with noisy, correlated, and unstructured pattern are bad. This visualization illustrates that the proposed SSF-CNN utilizes good filters. We next support these assertions with experimental results.

\vspace{6pt}

\noindent \textbf{Performance with Shallow Dictionary:}  To analyze the performance of the proposed method with varying training data sizes, 14 subsets of the training data of size $100,200,\cdots,1000,2000,\cdots, 5000$, are created. These sets are used to train the dictionary and SSF-CNN on each of the three databases individually. To train shallow dictionary for each database, 50 atoms are initialized and trained with varying number of training samples. The trained dictionary is then utilized to compute sparse features for training and testing samples. These features are input to a 3 layer neural network with 2 hidden layers of size $\{40, 20\}$. The results of shallow dictionary learning on three object classification databases are reported in Figure \ref{fg:db_graphs}. From these results, it can be inferred that shallow dictionary learning might not require large training data and increasing data may not lead to large improvement in classification results. This figure also shows that shallow dictionary learning may not be able to yield high classification accuracy and deep CNN architectures may further help.



\vspace{6pt}
\noindent \textbf{Performance with SSF-CNN and Comparison with Existing Algorithms:} We next evaluate the performance of the proposed SSF-CNN on three object classification databases by varying the training data size. The results in Figure \ref{fg:db_graphs} show that, in general, Xavier and MSRA initialization yield lower performance compared to the proposed dictionary initialization for very small training data. It can be consistently observed that the differences in results are more profound when the strength parameter \textbf{t} is learned with fixed \textbf{W}. The results further show that the performance of the proposed SSF-CNN increases with increase in training database size. It can be inferred that unlike shallow dictionary, where the performance does not improve significantly with increase in training database size, the parameters learned by the proposed SSF-CNN evolves with large data.

We also observe that the proposed algorithm, in general yields higher performance compared to three existing algorithms, PCANet~\cite{chan2015pcanet}, Deep Hybrid Network~\cite{oyallon2017scaling}, and ScatNet~\cite{anden2011multiscale}. We next perform the experiments when the structure of the filters are obtained from training on ImageNet data and then strength parameter is used to adapt to small sample size problem (i.e. Proposed ResNet: Pretrained on ImageNet, Learn $\mathbf{t}$). Results in Figure \ref{fg:db_graphs} show that our hypothesis that the structure of filters can be learned from training on large databases and knowledge can be \textit{adapted} with small training data using the strength $\mathbf{t}$ is valid.



\vspace{10pt}
\noindent \textbf{Results on Complete Training Data:} We have also evaluated the proposed dictionary learning based initialization method on the standard training protocols of all three databases, i.e., using the complete training data. Similar to small training data size, the experiments are performed with multiple methods of initializations and two ways of learning $\mathbf{W}$ and $\mathbf{t}$, i.e., (i) \textit{learn} $\mathbf{W}$ and (ii) \textit{learn} $\mathbf{t}$, \textit{freeze} $\mathbf{W}$. 
In this experiment, the proposed dictionary learning based initialization for ResNet is compared with Xavier and MSRA initialization. On the MNIST database, the proposed initialization yields an accuracy of $99.70$\% which is comparable with $99.71$\% achieved by standard initialization. On the NORB database, the proposed approach yields at least 3.8\% higher classification accuracy compared to existing initialization approaches. 
It is also observed that even if the filters have random values, learning strength produces considerably high accuracies. 
Once the filters are trained, optimizing the strength of those filters can further improve the performance. 

\subsection{Small Sample Size Case Studies}

\begin{table*}[]
\centering
\footnotesize
\caption{Rank-1 identification accuracies (\%) on the newborn face database \cite{bharadwaj2016domain}. The results are reported for fine-tuned pre-trained models and with learning the strength of pre-trained filters. The last three models are trained on face databases and the remaining models are trained on ImageNet~\cite{imagenet_cvpr09} database. }
\label{tb:newborn}
\begin{tabular}{|c||c|c|c|c||c|c|c|c|}
\hline
\multicolumn{1}{|c||}{\multirow{3}{*}{\textbf{\begin{tabular}[c]{@{}c@{}}Pre-trained \\ Model\end{tabular}}}} & \multicolumn{8}{c|}{\textbf{Number of Gallery Images}}                                                       \\ \cline{2-9} 
\multicolumn{1}{|c||}{}  & \multicolumn{4}{c||}{\textbf{Fine-tuning}}         & \multicolumn{4}{c|}{\textbf{Proposed Strength Learning}}   \\ \cline{2-9} 
\multicolumn{1}{|c||}{}   & \textbf{1} & \textbf{2} & \textbf{3} & \textbf{4} & \textbf{1} & \textbf{2} & \textbf{3} & \textbf{4} \\ \hline
ResNet 50                                                                                                    & 35.77 $\pm$  2.34   & 43.59$\pm$ 0.92      & 49.90 $\pm$  2.57   & 52.14    $\pm$ 3.31  & 37.80  $\pm$ 2.01    & 46.77$\pm$ 1.79      & 52.61 $\pm$  1.89   & 56.73 $\pm$ 1.79    \\ \hline
ResNet 101                                                                                                   & 35.86 $\pm$ 2.78    & 45.90$\pm$ 2.54     & 51.17$\pm$   2.10   & 54.59 $\pm$ 3.41   & 36.62 $\pm$  4.06   & 46.71 $\pm$ 3.73    & 52.79 $\pm$   1.72  & 56.16 $\pm$ 3.07     \\ \hline
ResNet152                                                                                                    & 36.30 $\pm$ 3.19   & 46.74 $\pm$ 2.42    & 51.99 $\pm$    2.24 & 55.47 $\pm$ 2.34 & 38.30 $\pm$  3.57   & 47.92 $\pm$  2.29   & 53.71 $\pm$ 2.62    & 59.57 $\pm$ 2.46    \\ \hline
VGG13                                                                                                        & 56.34 $\pm$ 2.46   & 68.49 $\pm$ 3.07     & 73.37 $\pm$ 2.53  & 76.47 $\pm$ 2.33    & 65.54  $\pm$ 3.20   & 78.14 $\pm$1.97     & 84.05 $\pm$  1.40    & 87.76 $\pm$1.88     \\ \hline
VGG16                                                                                                        & 57.07 $\pm$ 2.85   & 67.84 $\pm$ 2.61    & 73.21$\pm$ 3.10     & 76.21 $\pm$ 2.86    & 65.29 $\pm$ 1.99     & 79.18 $\pm$ 2.85    & 84.24  $\pm$ 2.82   & 87.50 $\pm$  1.47   \\ \hline
VGG19                                                                                                        & 53.87 $\pm$ 4.49   & 66.95 $\pm$ 2.15    & 72.33$\pm$ 1.25    & 75.75 $\pm$ 1.77    & 62.29 $\pm$1.70      & 75.36  $\pm$ 2.03   & 80.90  $\pm$0.77    & 84.20 $\pm$ 0.75    \\ \hline
DenseNet161                                                                                                  & 50.64 $\pm$ 3.27    & 63.65 $\pm$ 2.95     & 68.98$\pm$ 1.79    & 72.86     $\pm$1.82  & 58.39 $\pm$5.59     & 72.14 $\pm$1.82     & 77.36 $\pm$ 1.57    & 81.04  $\pm$ 1.40   \\ \hline
DenseNet169                                                                                                  & 54.15$\pm$ 4.33     & 68.91$\pm$  2.99    & 73.31$\pm$    1.72 & 72.97 $\pm$ 2.05     & 58.25 $\pm$ 1.68     & 73.10 $\pm$0.99     & 78.91 $\pm$ 1.02    & 83.31 $\pm$ 1.12    \\ \hline
DenseNet201                                                                                                  & 60.78 $\pm$2.00     & 71.19 $\pm$ 0.84     & 71.48$\pm$2.17      & 73.64  $\pm$1.39    & 61.45  $\pm$  5.09  & 74.58 $\pm$2.40     & 80.75  $\pm$  3.86  & 85.02  $\pm$3.98    \\ \hline
LightCNN-9                                                                                                   & 55.72  $\pm$2.90    & 66.09 $\pm$2.27     & 67.65 $\pm$2.29     & 71.81 $\pm$1.64    & 56.48 $\pm$ 4.60   & 69.82 $\pm$4.49    & 76.91$\pm$3.69     & 81.87$\pm$ 3.93    \\ \hline
LightCNN-29                                                                                                  & 53.10 $\pm$3.75     & 65.28 $\pm$2.47     & 71.91$\pm$1.99 & 75.85 $\pm$2.02     & 62.67 $\pm$2.59     & 76.19  $\pm$1.15    & 82.55 $\pm$ 0.87    & 86.00 $\pm$ 1.03   \\ \hline
VGG-Face                                                                                                     & 60.77 $\pm$1.28     & 72.93 $\pm$1.40    & 77.19 $\pm$1.27   & 79.66 $\pm$1.97     & \textbf{70.42 $\pm$0.50}      & \textbf{81.37 $\pm$1.59}    & \textbf{86.50 $\pm$1.20}     & \textbf{90.01$\pm$1.53}  \\ \hline
\end{tabular}
\end{table*}

\begin{figure}[!t]
	\centering
	\includegraphics[width=.425\textwidth]{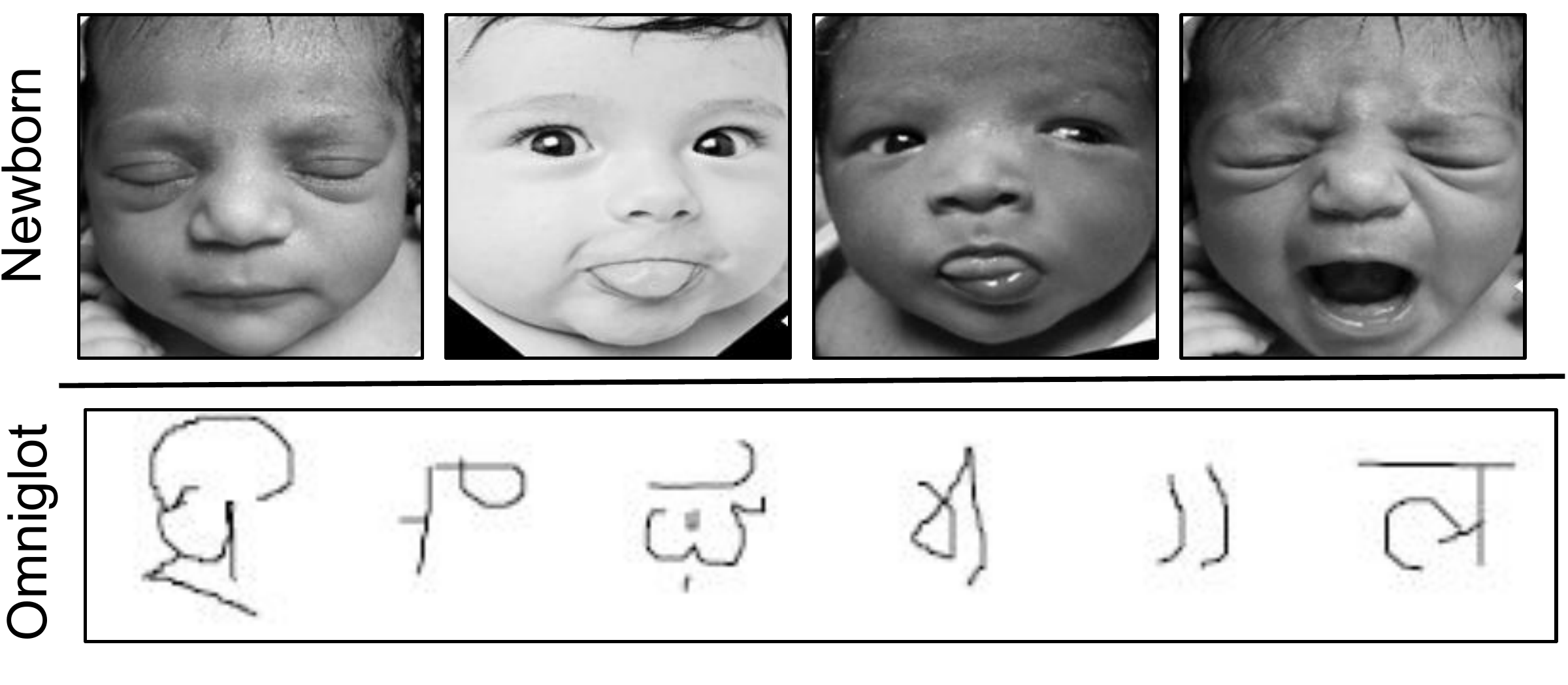}
	\caption{Samples images from the Omniglot and Newborn Faces databases.}	
	\label{fig:db_image} 	
\end{figure}




To showcase the effectiveness of the proposed \textit{structure} and \textit{strength} concept on small sample size databases, we present two case studies (i) newborn face database \cite{bharadwaj2016domain} and (ii) Omniglot database \cite{lake2011one}. Figure \ref{fig:db_image} shows sample images from both the databases.

\vspace{6pt}
\noindent \textbf{Newborn Face Recognition}: Bharadwaj \textit{et al.} \cite{bharadwaj2016domain} have shown that newborn face recognition is a challenging small sample size application. The publicly available IIITD Newborn database \cite{bharadwaj2016domain} contains face images from 96 newborns. The pre-defined protocol limits us to use training samples from only 10 newborns and testing is performed with 86 newborns. We compute the performance of ResNet architecture where the proposed dictionary based initialization helps in estimating the structure using images from 10 newborns and then strength parameter is used to attune the filters. The observed rank-1 accuracy in this case is 36.32\% which is at least 0.5\% better than pre-trained ResNet architecture (which is traditionally fine-tuned with newborn training data). Also, when we use training images of only 10 newborns to train filter of CNN models from scratch, the test accuracies are extremely low.

\begin{figure}[!t]
	\centering
	\includegraphics[width=0.5\textwidth]{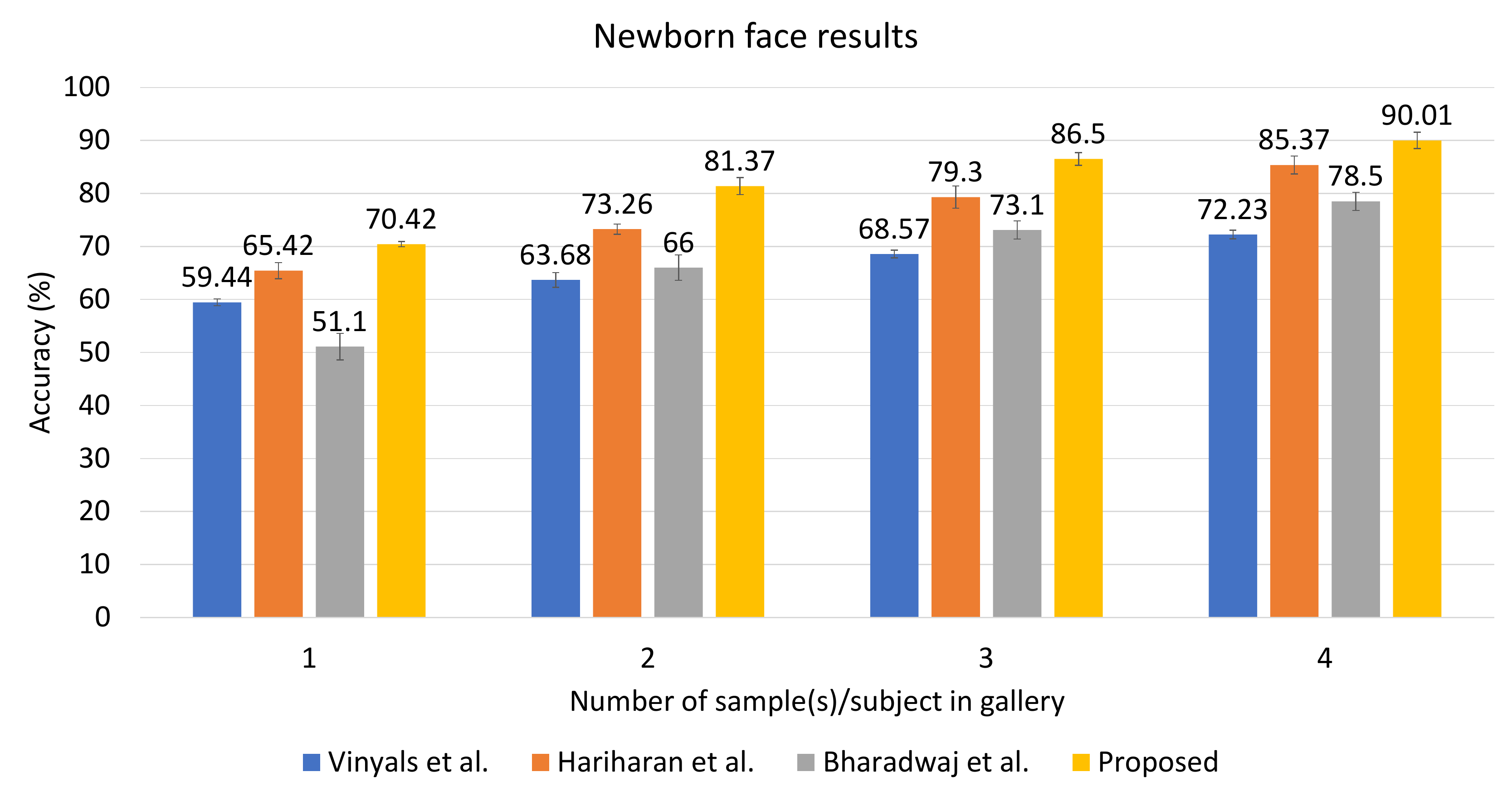}
	\caption{Summarizing the results on the newborn face database. }
	\label{fig:soa}
\end{figure}

As discussed before, we can learn ``structure'' from large domain-specific data and then the proposed ``strength'' can help attune the filters for problem-specific data. Therefore, we perform experiments with pre-trained networks (pre-trained filters are obtained after learning from either ImageNet or Labeled Faces in the Wild dataset (LFW)~\cite{huang2007labeled} and YouTube Faces (YTF)~\cite{wolf2011face} databases) and use strength parameter to attune it for newborn face recognition based on training data of 10 newborns. For this experiment, as shown in Table~\ref{tb:newborn}, we use variants of ResNet~\cite{he2016deep},  VGG~\cite{simonyan2014very}, VGGFace~\cite{Parkhi15}, LightCNN~\cite{wulight}, and DenseNet~\cite{huang2017densely} architectures, and the performance is compared with standard fine-tuning approaches using same images from 10 newborns.  
As shown in Table~\ref{tb:newborn}, we have observed that learning strength of the filters improves the performance of CNN models compared to conventional fine-tuning approach. With single gallery image per subject, the best rank-1 accuracy of over 70\% is obtained when the proposed strength parameter is used with pre-trained VGG-Face~\cite{Parkhi15} which is at least 10\% better than the conventional fine-tuning based approach. This shows that in real-world applications, the concept of learning structure and strength helps in achieving improved performance. 


The performance of the proposed approach is also compared with deep hybrid network~\cite{oyallon2017scaling} and ScatNet~\cite{anden2011multiscale}. For one gallery per subject, the rank-1 accuracies of these two algorithms are $25.18\%\pm1.33\%$ and $31.04\%\pm1.94\%$ respectively, which are at least 39\% less than the best results reported in Table~\ref{tb:newborn}. Finally, we also compare the performance of the proposed algorithm with the Vinyals~\textit{et al.}~\cite{vinyals2016matching}, Hariharan~\textit{et al.}~\cite{hariharan2016low}, and Bharadwaj~\textit{et al.}~\cite{bharadwaj2016domain} on newborn face database. Using the same protocol, Figure~\ref{fig:soa} illustrates the comparison between the proposed method (best reported result in Table~\ref{tb:newborn}) with existing methods. The proposed method improves the rank-1 accuracies by $11 - 19\%$ for varying number of sample(s) per subject. However, the proposed algorithm consistently yields improved accuracies and is approximately 4.5\% better than the second best performing approach~\cite{hariharan2016low}. 

\vspace{6pt}

\noindent \textbf{Omniglot Database}: On the Omniglot database \cite{lake2011one}, SSF-CNN yields classification accuracies of $97.6\%\pm0.84\%$ and $98.3\%\pm1.03\%$ for 1-shot, 5-way and 5-shot, 5-way, respectively which are comparable to state of the art results. Table \ref{tab:omniglot} summarizes the results of the proposed algorithm and compares them with existing algorithms. The results show that SSF-CNN is among the top performing algorithms for both the protocols.

\section{Discussion and Conclusion}

Large training database is a key requirement for training convolutional neural networks. However, there are several applications and problem statements that do not have the luxury of large training databases. In this research, we propose Structure and Strength Filtered CNN as a framework for learning a CNN model with small training databases. We propose to initialize the filters of CNN using dictionary filters which can be trained with small training samples. Since the dictionary atoms are learned for reconstruction, they may not be optimal for classification. Therefore, we next suggest to learn the strength of the filters with the given training data. The effectiveness of the proposed model has been demonstrated on multiple object classification databases and a real-world newborn face recognition problem. Using different architectures and experiments, we demonstrate the efficacy of the proposed approach. Specifically, in case of newborn face recognition, remarkable improvement in accuracy is achieved with the proposed approach. The proposed CNN has the flexibility to work for small as well as large databases. The current model incorporates unsupervised dictionary filters to initialize the CNN network. As a future work, other trained filters such as supervised dictionary filters can also be used. They can also be used to adapt the filters from one task to another task while learning only the strength of the filters. The proposed algorithm can also be extended to other applications such as face recognition with variations in disguise \cite{disguise}, matching faces in videos \cite{ggvideo}, and sketch to photo matching \cite{sketch}.

\begin{table}[]
\centering
\caption{Classification results (\%) on the Omniglot database~\cite{lake2011one}.}
\label{tab:omniglot}
\begin{tabular}{|c|c|c|}
\hline
\textbf{Algorithm}                                                        & \textbf{1-shot, 5-way} & \textbf{5-shot, 5-way} \\ \hline
Santoro~\textit{et al.}~\cite{santoro2016meta}     & 82.8                   & 94.9                   \\ \hline
Koch~\textit{et al.}~\cite{koch2015siamese}        & 97.3                   & 98.4                   \\ \hline
Vinyals~\textit{et al.}~\cite{vinyals2016matching} & 98.1                   & 98.9                   \\ \hline
\textbf{Proposed}                                                         & 97.6             & 98.3             \\ \hline
\end{tabular}
\end{table}

\section{Acknowledgment}

This research is partially supported by Ministry of Electronics and Information Technology, India. Rohit Keshari is partially supported by Visvesvaraya Ph.D. fellowship. Richa Singh and Mayank Vatsa are partly supported by the Infosys Center of Artificial Intelligence, IIIT Delhi, India.




{\small
\bibliographystyle{ieee}
\bibliography{cnn_on_small}
}

\end{document}